# Vehicle-group-based Crash Risk Prediction and Interpretation on Highways

Tianheng Zhu, Ling Wang, Yiheng Feng, *Member, IEEE*, Wanjing Ma and Mohamed Abdel-Aty, *Senior Member, IEEE*

***Abstract*—Previous studies in predicting crash risks primarily associated the number or likelihood of crashes on a road segment with traffic parameters or geometric characteristics, usually neglecting the impact of vehicles' continuous movement and interactions with nearby vehicles. Recent technology advances, such as Connected and Automated Vehicles (CAVs) and Unmanned Aerial Vehicles (UAVs), are able to collect high-resolution trajectory data, which enable trajectory-based risk analysis. This study investigates a new vehicle group (VG) based risk analysis method and explores risk evolution mechanisms considering VG features. An impact-based vehicle grouping method is proposed to cluster vehicles into VGs by evaluating their responses to the erratic behaviors of nearby vehicles. The risk of a VG is aggregated based on the risk between each vehicle pair in the VG, measured by inverse Time-to-Collision (iTTC). Logistic Regression and a Graph Neural Network (GNN) are used to predict VG risks based on both aggregated and disaggregated VG information. Both methods achieve excellent performance with AUC values exceeding 0.93. For the GNN model, GNNExplainer with feature perturbation is applied to identify critical individual vehicle features and their directional impact on VG risks. Overall, this research contributes a new perspective for identifying, predicting, and interpreting traffic risks.**

This study is supported by National Natural Science Foundation of China (No. 52325210, 52372333, 52202420) and Fundamental Research Funds for the Central Universities (2023-4-YB-05). (Corresponding author: Ling Wang.)

Tianheng Zhu was with the College of Transportation Engineering, Tongji University, Shanghai, 201804 China. He is now with the Lyles School of Civil and Construction Engineering, Purdue University, Indiana, 47907 USA (e-mail: zhu1230@purdue.edu).

Ling Wang is with the College of Transportation Engineering, Tongji University, Shanghai, 201804 China (e-mail: wang_ling@tongji.edu.cn).

Yiheng Feng is with the Lyles School of Civil and Construction Engineering, Purdue University, Indiana, 47907 USA (e-mail: feng333@purdue.edu).

Wanjing Ma is with the College of Transportation Engineering, Tongji University, Shanghai, 201804 China (e-mail: mawanjing@tongji.edu.cn).

Mohamed Abdel-Aty is with the Department of Civil, Environmental and Construction Engineering, University of Central Florida, Florida, 32816 USA (e-mail: M.Aty@ucf.edu).

## I. INTRODUCTION

The evolution of communication, sensing and computing technologies brings new opportunities to Proactive Traffic Safety Management (PTSM) systems. With Vehicle-to-Everything (V2X) communication, Connected and Automated Vehicles (CAVs) not only provide a new source of data, but also can be controlled to mitigate risks and enhance traffic safety. To implement CAV-based PTSM applications, such as collision warning and avoidance, a critical pre-step is to identify and predict the crash risks between CAVs and their neighboring vehicles.

There have been plenty of studies on crash risks for highways or expressways. Most of them focused on road segments as their subject of analysis, with potential crash predictors extracted from the current, upstream, and downstream segments [1-4]. Meanwhile, the crash-contributing factors are aggregated at the segment level, e.g., traffic flow features [5], roadway geometric characteristics [1], etc. However, due to vehicles' continuous movement, the vehicle state at one point can influence future risks. Meanwhile, crash risks often result from interactions between vehicles and evolve over time. The vehicle dynamics and their interactive behaviors are critical to the understanding of crash risks. Wang et al. [6] attempted to consider vehicle dynamics in risk analysis by deriving historical vehicle positions through space-mean-speed and constructing a quasi-vehicle trajectory. Traffic data collected from the road segments of these positions were utilized to understand crash risks. However, the input data collected from fixed traffic sensors were still at the aggregated level, which lack details on individual vehicles and the interactions between them. With the advancement of V2X communication technology and vehicle onboard sensors, the presence of other vehicles in the immediate neighborhood can be observed [7], which provide detailed vehicle trajectory data. This new data source allows the prediction and analysis of crash risks from the perspective of a group of vehicles. Existing studies have considered vehicles as groups and conducted related research in this domain. For example, Li et al. [8] incorporated the concept of safety potential field to describe the driving risks of vehicles in platoons. He et al. [9] developed an integrated variable speed limit and ramp metering control framework that includes a vehicle group risk prediction model aiming at reducing crash and congestion risks of vehicle groups. However, to the best of our knowledge, few studies have shed light on the interpretation of the mechanisms underlying the vehicle-group (VG) based risk formation and evolution.

Methodologically, crash risk prediction methods can be broadly categorized into three types: statistical, machine learning (ML), and deep learning (DL) approaches. Statistical and traditional ML methods, such as Logistic Regression [4, 6] and Decision Tree [10], are valued for their simplicity, interpretability, and ability to model relationships in data effectively. Deep learning approaches, including Convolutional Neural Network (CNN) [11], Recurrent Neural Network (RNN) [12, 13], and Transformers [14, 15], excel at capturing complex, non-linear relationships and patterns in large, high-dimensional datasets. For example, Yuan et al. [12] predicted real-time crash risk by considering time series dependency with the employment of a long short-term memory recurrent neural network (LSTM-RNN) algorithm. Han et al. [15] employed a Transformer model with self-attention mechanism to evaluate traffic risks using edge scenarios such as hard braking, sudden acceleration, and aggressive lane change. However, traditional DL methods struggle to handle the irregularity and complexity of graph-structured data (e.g., a group of interacting vehicles). Graph Neural Network (GNN) [16] has been proven to be effective in modeling spatiotemporal data and widely implemented in many traffic applications, such as traffic forecasting [17, 18] and risk assessment [19]. In this study, a GNN architecture is employed to predict risks using the graph data extracted from groups of interacting vehicles.

Interpretation of the model result allows stakeholders to understand why the model produces such a prediction, which is crucial in model improvement, trustworthiness and practical implications. Many statistical and ML models, like Logistic Regression [4, 6], Classification and Regression Tree (CART) [20, 21], and Multivariate Adaptive Regression Splines (MARS) [22, 23], are naturally explainable [24]. However, the black-box nature of DL models, including the GNN model used in this study, require post-hoc analysis, or “Model-Agnostic Method” [25] for better interpretation. Several post-hoc analysis methods have been proposed, including saliency map [26], DeepLIFT [27] or DeepSHAP [28]. However, saliency map may be misleading in some instances [29]. Given the dynamic nature of VGs, it’s challenging to define an appropriate baseline for DeepLIFT or DeepSHAP. These approaches also fall short in their ability to incorporate relational information, which is the essence of graphs [30]. Therefore, GNNExplainer [30] is applied in this study, which is capable of identifying rich subgraphs and important node features of graph inputs in a concise and consistent way. Feature perturbation is also applied to the identified critical node features to reveal their directional impact on the prediction results.

The main contributions of this paper are as follows:

(1) We propose an impact-based vehicle grouping method to form VGs so that the interactive nature of crash risks can be captured.

(2) We employ an Edge-Conditioned Convolutional (ECC) Graph Neural Network to predict crash risks using graph data of VGs, where nodes represent individual vehicles and edges represent the risks between vehicles.

(3) To provide better interpretation, we introduce an explainable framework that integrates GNNExplainer with feature perturbation to identify critical individual vehicle features with their directional impact on risks within VGs.

This paper is organized into six sections. Following this introduction, **Section II** details the methodologies, including vehicle grouping and association methods, VG risk quantification, and the prediction and interpretation models. **Section III** presents the data preparation process. **Section IV** presents the results on the prediction and interpretation of VG risks. Finally, **Section V** summarizes the conclusions, applications, and limitations of the study.

## II. Methods

The flow of this work is illustrated in **Figure 1**. We start with the vehicle grouping and the association of VGs in consecutive timestamps. After quantifying the VG risks, the dependent variables of the prediction models are defined and calculated. Using the aggregated and disaggregated VG information, two prediction models based on Logistic Regression and a Graph Neural Network are developed to predict two aspects of VG risk evolution (i.e., vehicle group risk level and trend of risk scale changes). Finally, model-based and post-hoc interpretation methods are applied to understand the factors influencing VG risks.

### *A. Definition of Vehicle Groups (VGs) and Vehicle-group Risk Calculation*

In this study, VGs are selected as the primary subjects to capture the dynamic and interactive nature of crash risks. We propose an impact-based vehicle grouping (IVG) method to form VGs based on the Intelligent Driver Model (IDM) [31]. Then, two association rules are applied to match the VGs in consecutive timestamps to understand the impact of current VG states on future risks.

#### *1) Impact-based Vehicle Grouping (IVG) method*

IVG groups vehicles based on the ego vehicle's reactions to the erratic behaviors of surrounding vehicles. This method quantifies the impact of surrounding vehicles by the deceleration rates of the ego vehicle after responding to vehicles from both the same lane and adjacent lanes.

For vehicles traveling in the same lane, we model a sudden brake of the leading vehicle with a predefined deceleration rate. Applying the IDM, the required deceleration rate of the ego vehicle is then calculated. If the response deceleration rate exceeds a predefined threshold, the leading and the ego vehicle are grouped together. The determination of the deceleration rate of the leading vehicle and the response deceleration rate threshold is discussed in **Section III.B**.

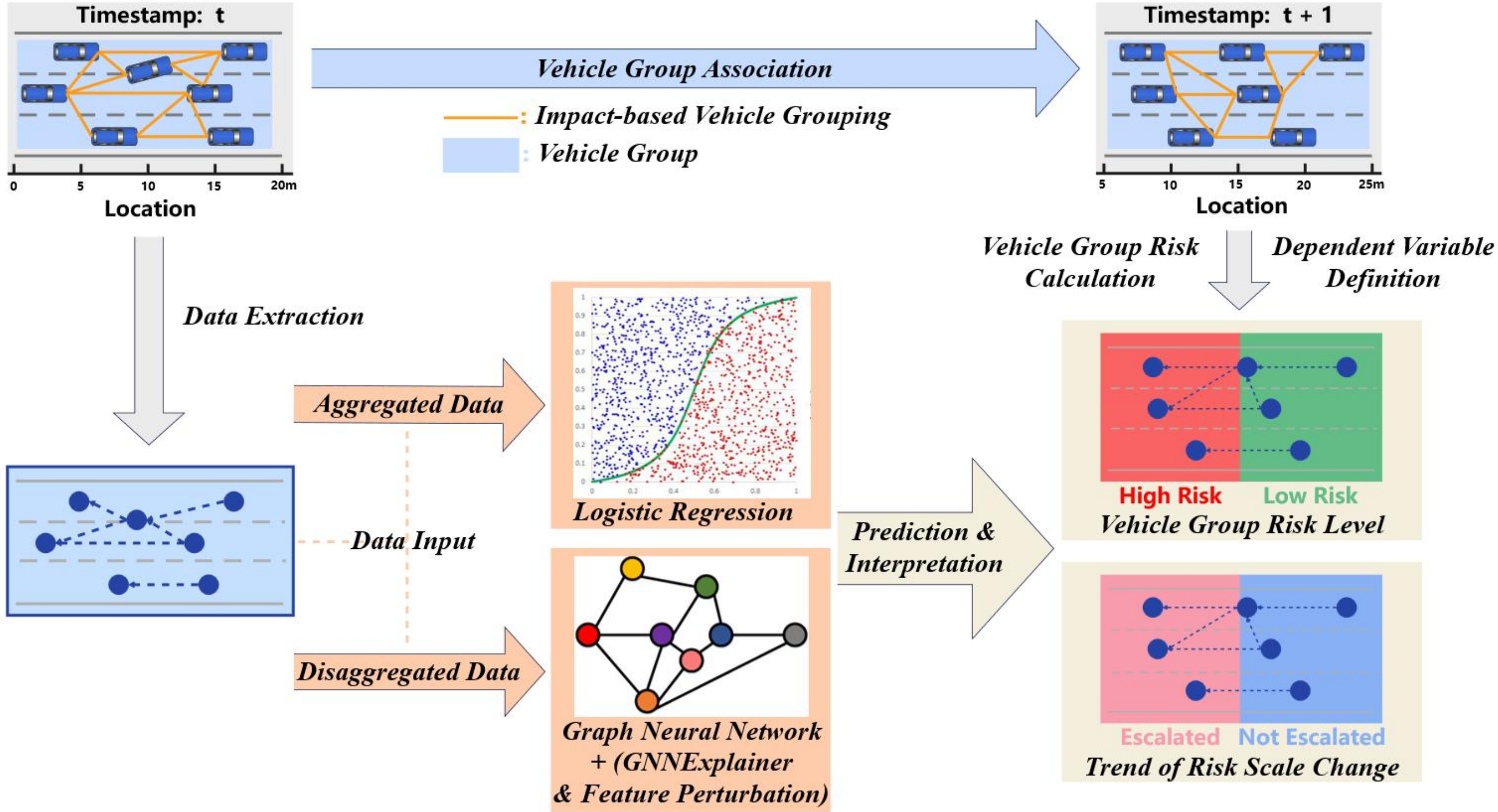

Fig. 1. Methodology Flowchart

For adjacent lanes, we employ a simplified constant velocity and constant angular velocity model [32], expressed in **Eq. (1),** to describe the vehicles' lane change behaviors**.** Then, we model a cut-in scenario where the nearest leading vehicles in the adjacent lanes perform an instantaneous lane change with a predefined heading angle (as discussed in **Section III.B**). The heading angle is used to measure the vehicles' speed after lane changes. The ego vehicle's deceleration rate after the cut-in of adjacent vehicles is calculated and compared against the same response deceleration rate threshold. If the deceleration exceeds the threshold, the ego vehicle and the adjacent vehicles are grouped together.

$$\dot{x} = v_x(t) = v_0 \cos(\varphi_0) \qquad (1)$$
$$\dot{y} = v_y(t) = v_0 \sin(\varphi_0)$$

where $\dot{x}$ and $\dot{y}$ are the first derivative of vehicles' lateral and longitudinal positions. $v_0$ and $\varphi_0$ are the initial vehicle speed and heading angle at the start of lane changes.

*2) Association of VGs at consecutive timestamps*

To understand the impact of current VG states on future risks, we associate the VGs across consecutive timestamps to establish the trajectories of VGs. Owing to random vehicle movements, VGs may differ across timestamps. To streamline the process, two association rules are applied:

- If one or more leading vehicles are shared between two VGs in consecutive timestamps, these groups are matched accordingly. The actions of leading vehicles significantly influence the behaviors of following vehicles. For example, Wang et al. [33] showed that the deceleration of the leading vehicle in a platoon led to a narrowing spacing between vehicles and the decrease of the overall speed of the platoon. Thus, the leading vehicles can effectively characterize a VG.
- When a VG has several matchable counterparts in the next timestamp following the first rule, the counterpart with the highest ratio of shared vehicles is selected as the most suitable match.

*3) VG risk calculation*

The risk of a VG is considered as the aggregation of the risks between all the leading and following vehicle pairs inside the VG. During calculation, specifically, the lane-changing vehicles are projected onto both their original and target lanes and treated as two vehicles to account for their influence on both lanes. In rare instances where the projected lane-changing vehicles overlap with other existing vehicles in the target lane, a high risk value (e.g., the 98$^{th}$ percentile of all risk values) is assigned to the affected vehicle pairs.

The risk between a leading and following vehicle pair is quantified using Surrogate Safety Measures (SSMs), an alternative method of assessing safety that relies on the analysis of safety-critical events known as traffic conflicts [34]. In this study, inverse Time-to-Collision (iTTC) [35], expressed in **Eq. (2)**, is adopted as the SSM to quantify crash risks. iTTC is defined as the inverse of the remaining time until a collision occurs, assuming both the leading and following vehicles maintain their current speeds and trajectories. It is chosen to address the issue of infinite values in the original Time-to-Collision (TTC) calculation, which arise when the following vehicle's velocity is lower than the leading vehicle, making computation feasible.

$$iTTC_{i,k}(j) = \begin{cases} \dfrac{v_{i,k}(j) - v_{i-1,k}(j)}{x_{i-1,k}(j) - x_{i,k}(j) - d_{i-1,k}}, & v_{i,k}(j) > v_{i-1,k}(j) \\ 0, & v_{i,k}(j) \le v_{i-1,k}(j) \end{cases} \quad (2)$$

where $iTTC_{i,k}(j)$ represents the inverse Time-to-Collision value of the $i^{\text{th}}$ vehicle and $i-1^{\text{th}}$ vehicle in the $k^{\text{th}}$ vehicle group at $j^{\text{th}}$ timestamp. $x_{i,k}(j)$ is the position of the vehicle. $d_{i,k}$ is the length of the vehicle. $v_{i,k}(j)$ is the speed of the vehicle.

Using iTTC, VG risk is aggregated based on two fronts:

- Risk Indicator. The Risk Indicator of a VG is binary and determined by the maximum risk value inside the group. By comparing against a threshold (iTTC = 2/3 $s^{-1}$, equivalent to TTC of 1.5s [36-38]), Risk Indicator is used to determine in general whether the VG is of high-risk or not.
- Risk Scale. The Risk Scale of a VG is defined as the number of the vehicle pairs with risk values exceeding the same threshold (iTTC = 2/3 $s^{-1}$) in the group. It is introduced to address the insufficiency of Risk Indicator in quantifying the VG risks. For example, even if two VGs are both identified to be high-risk using Risk Indicator, they may differ significantly in the number of hazardous interactions between vehicle pairs within VGs, leading to different implications for safety. Moreover, Risk Scale provides a complementary measure to study the evolution of VG risks.

### *B. VG Risk Prediction and Interpretation*

We develop two VG risk prediction models including Logistic Regression (LR) and a Graph Neural Network (GNN) using aggregated and disaggregated data respectively to predict two aspects of VG risk evolution: (1) the Vehicle Group's Risk Level (VGRL) at the next timestamp (high-risk or not); and (2) the Trend of Risk Scale Change (TRSC) within a VG at the next timestamp (escalated or not). Both dependent variables are binary. As defined in **Section II.A.(3)**, a VG is classified as high-risk (label = 1) if its Risk Indicator exceeds the threshold of 2/3 $s^{-1}$; otherwise, it is considered as low risk (label = 0). TRSC is determined by comparing the values of Risk Scale in consecutive timestamps: if the Risk Scale is escalated, the TRSC is 1 and 0 otherwise. The predictions of the two dependent variables are conducted separately.

#### *1) Logistic Regression*

LR, as outlined in **Eq. (3)**, is a robust classification model, offering a clear interpretation of relationships between independent and dependent variables.

$$log\frac{p}{1-p} = \beta_0 + \beta_1 x_1 + \beta_2 x_2 + \cdots + \beta_m x_m \quad (3)$$

where $p$ is the probability of classifying as Class 1. $\beta_0$ is the intercept. $\beta_i$, $i$ =1, 2, …, m, is the coefficient of independent variable $x_i$.

The aggregated VG information (the independent variables of the model) encompasses the vehicle motions and types in VGs, alongside their current risk and relationship with road junctions, as detailed in **Table I**.

#### *2)* Graph Neural Networks

A VG can be represented as a directed graph $G = (V, E)$, where $V$ is the set of nodes and $E \in V \times V$ is the set of Edges. The nodes represent individual vehicles in a VG, while the edges denote the relationships between two vehicles. This disaggregated data enables the analysis of the impact of individual vehicles on VG risks.

Each node is assigned eight features capturing the vehicle's position, driving information, type, and relationship with road junctions, while the risk between two vehicles serves as the sole edge attribute. All features are detailed in **Table II**.

The selection of GNN architecture follows two guiding principles:

- Enhancing Interpretability: To ensure the interpretability of the impact of node features, the path of information propagation within the network should remain relatively straightforward. Consequently, GNN architectures employing attention mechanisms, such as Graph Attention Networks (GATs) [39] and Graph Transformer Networks (GTNs) [40], are excluded. While these models are powerful, the inclusion of attention weights introduces additional complexity.
- Incorporating Edge Attribute: The edge attribute, specifically the risk between vehicles, is a key predictor of VG risks. Therefore, the GNN architectures that do not explicitly incorporate edge attributes but instead focus on node features and structural connections, like Dynamic Graph Convolutional Neural Network (DGCNN) [41] and GraphSAGE (SAmple and aggreGatE) [42], are deemed unsuitable.

Following the two rules, Edge-Conditioned Convolutional (ECC) Graph Neural Network [43], which has demonstrated strong performance in graph classification tasks, is chosen for this study. ECC employs edge-conditioned convolutions to dynamically generate filter weights based on edge attributes through learnable networks, as described in **Eq. (4)**. This approach enables seamless integration of edge attributes into the model's reasoning process. An illustration of the ECC network used in this study is shown in **Figure 2**.

$$\begin{aligned} X^l(i) &= \frac{1}{|N(i)|} \sum_{j \in N(i)} F^l(L(j,i); w^l) X^{l-1}(j) + b^l \\ &= \frac{1}{|N(i)|} \sum_{j \in N(i)} \Theta_{ji}^l X^{l-1}(j) + b^l \end{aligned} \quad (4)$$

where $d_l$ and $d_{l-1}$ are the feature dimensions of node $i$ at layer $l$ and layer $l-1$. $X^l(i) \in R^{d_l}$ is the filtered signal at node $i$ at layer $l$, which is computed as a weighted sum of signals $X^{l-1}(i) \in R^{d_{l-1}}$ in its neighborhood, $j \in N(i)$. $F^l$ is the filter-generating network (a Multilayer Perceptron (MLP) in our study), parameterized by learnable network weights $w^l \in R^{d_l}$, which outputs edge-specific weight matrix $\Theta_{ji}^l \in R^{d_l \times d_{l-1}}$ given edge attribute $L(j,i)$. $b^l \in R^{d_l}$ is a learnable bias. For clarity, $w^l$ and $b^l$ are model parameters updated only during training and $\Theta_{ji}^l$ are dynamically generated parameters for the edge attributes in a particular input graph.

*3) Explainable GNN Framework*

While GNNs excel at capturing intricate patterns in graph-structured data, their decision-making processes are often opaque, making it challenging to interpret how specific nodes, or edges contribute to predictions. This challenge is further amplified by the heterogeneity of VG compositions (i.e., the number of vehicles and their relations in a VG evolve over time), resulting in the dynamic graph structures. To address these issues, we introduce an explainable framework that combines GNNExplainer with feature perturbation to interpret GNN results. This framework identifies critical individual vehicle features and evaluates their directional impact on VG risks, offering valuable insights into the factors driving GNN predictions in dynamic graph structures.

GNNExplainer [30] is a powerful tool designed to interpret and explain the predictions made by GNN. It analyzes a trained GNN along with its predictions to produce an explanation in the form of a rich subgraph of the input graph, accompanied by a subset of node features. The subgraph is selected to maximize the mutual information with the GNN's predictions. This is achieved by formulating a mean-field variational approximation and learning a real-valued graph mask. In our case, the graph mask explicitly identifies the critical vehicles within a VG that are most influential to VG risks, along with their associated features.

Then feature perturbation is applied to the identified significant features to evaluate how changes in these features influence the VG risks. This technique can be utilized to assess the directional impact of features by systematically reducing their magnitude and observing the effect on the output. If reducing the magnitude of a node feature leads to a corresponding decrease in the output, the feature is deemed to have a positive impact. Conversely, if increase in output is observed, the feature is considered to have a negative impact. This approach provides a straightforward way to quantify the contribution of individual node features to the model's predictions.

TABLE I
THE INDEPENDENT VARIABLES FOR LOGISTIC REGRESSION

| Variables | Definition | Unit |
|---|---|---|
| *avg_v* | the average (*avg*) velocities (*v*) of vehicles within a VG | m/s |
| *avg_a* | the average (*avg*) absolute acceleration rates (a) of vehicles within a VG | |
| *std_v* | the standard deviation (*std*) of vehicle velocities (*v*) within a VG | m/s |
| *pctg_large_veh* | the percentage (*pctg*) of large vehicles (*large_veh*), i.e., heavy car (10m) and bus (12m) in a VG | % |
| *pctg_change_lane* | the percentage (*pctg*) of vehicles executing lane changes (*change_lane*) in a VG | % |
| *risk_indicator* | the current Risk Indicator of a VG, as defined in **Section II.A.(3)** | $s^{-1}$ |
| *num_on_ramp* | the number (*num*) of vehicles within a 100-meter range of the on-ramp (*on_ramp*) in a VG | - |
| *num_off_ramp* | the number (*num*) of vehicles within a 100-meter range of the off-ramp (*off_ramp*) in a VG | - |

TABLE II
DEFINITION OF NODE FEATURES AND EDGE ATTRIBUTE IN GRAPHS

| Features | Node/Edge | Definition | Unit |
|---|---|---|---|
| *lane_id* | node | the index of lane (*lane_id*) where a vehicle is driving on | - |
| *dist* | node | the relative distance (*dist*) of a vehicle measured along the centerline from the starting point of the highway | m |
| *v* | node | velocity (*v*) of a vehicle | m/s |
| *abs_a* | node | absolute value (*abs*) of acceleration (*a*) of a vehicle | $m/s^2$ |
| *type* | node | type (*type*) of the vehicle, i.e., car, medium car, heavy car, bus | - |
| *change_lane* | node | whether a vehicle is changing lanes | binary |
| *on_ramp* | node | whether a vehicle is within a 100-meter range of an on-ramp (*on_ramp*) | binary |
| *off_ramp* | node | whether a vehicle is within a 100-meter range of an off-ramp (*off_ramp*) | binary |
| *risk* | edge | the risk between two vehicles measured by iTTC, as defined in **Section II.A.(3)** | $s^{-1}$ |

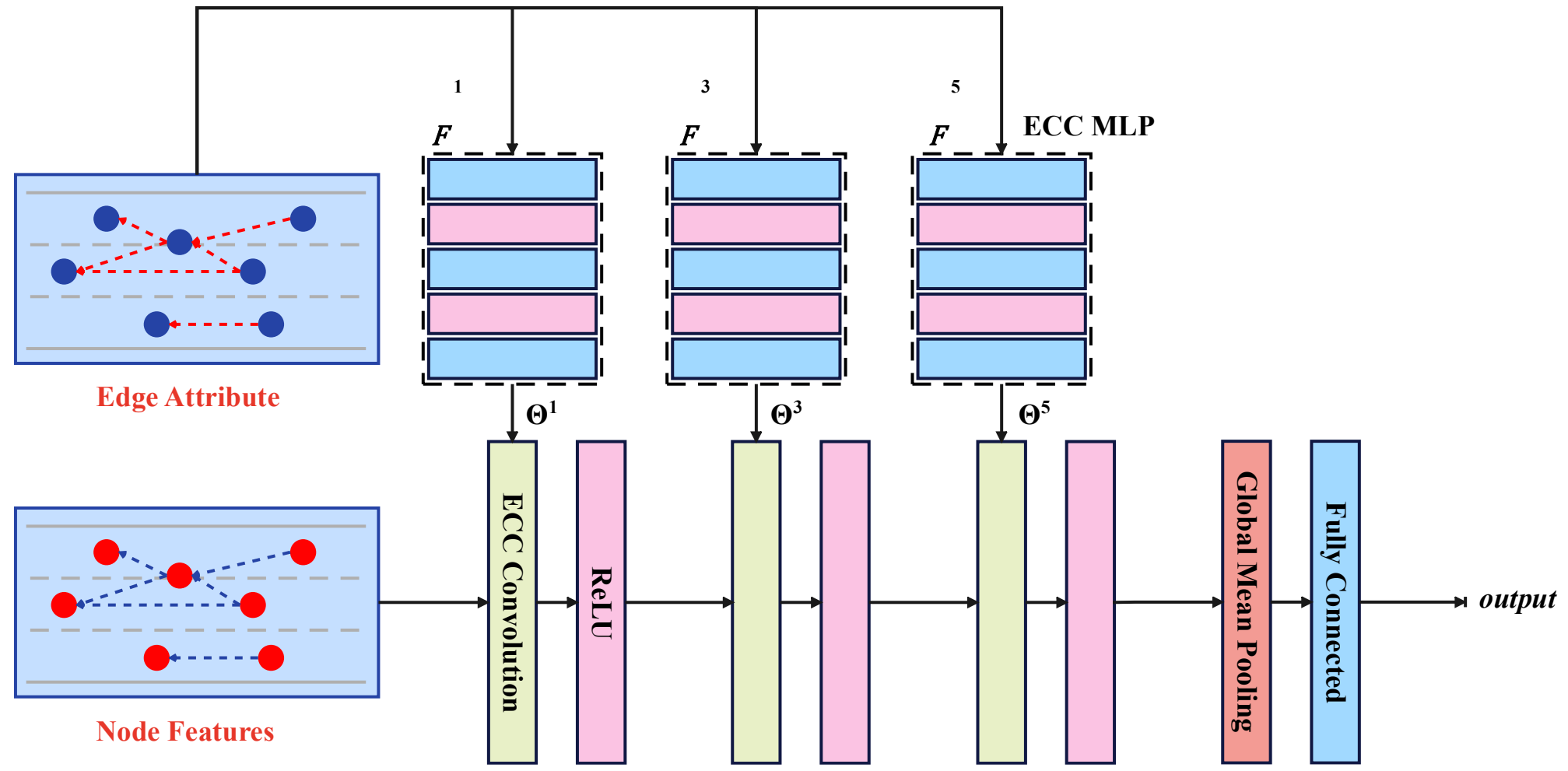

Fig. 2. Illustration of ECC architecture

## III. DATA PREPARATION

### A. Trajectory Dataset

The trajectory dataset used in this study is the MAGIC Dataset [44], with the locations of the road sections illustrated in **Figure 3**. To collect the data, multiple Unmanned Aerial Vehicles (UAVs) were used to capture about 3 hours of aerial video at six locations along a 4,000 m urban highway. The vehicle trajectories were then extracted with a frequency of 0.04 s (i.e., 25 Hz). The dataset includes vehicle ID, type, position, speed, acceleration, timestamp, lane ID (a decimal value, such as 1.5, indicates a lane change between lane 1 and lane 2), and direction. For consistency, only the eastbound data is used in the analysis.

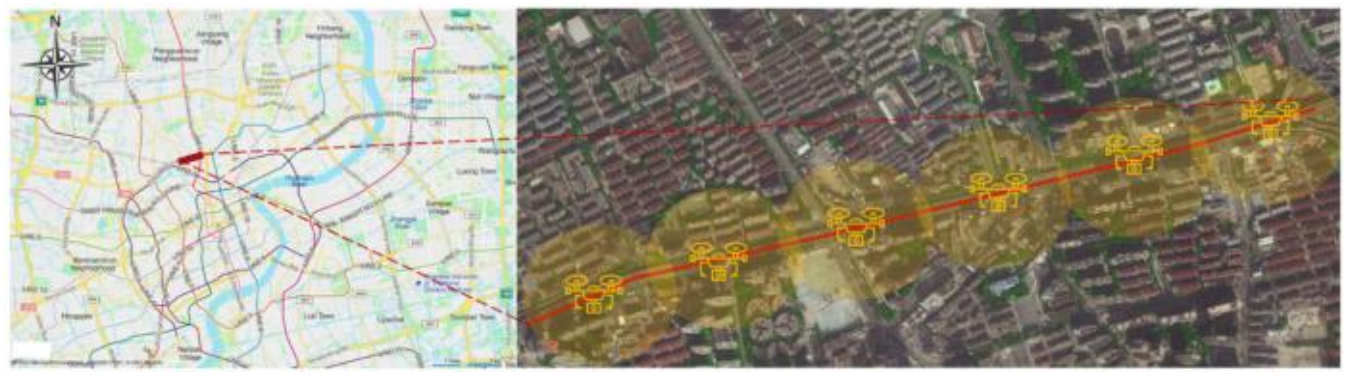

Fig. 3. Location of the road section

### B. Model Calibration and Parameter Determination

To implement the vehicle grouping and association methods introduced in **Section II.A**, the IDM model and a few critical parameters need to be determined first.

The IDM, expressed in **Eq. (5)**, is calibrated using data from 100 vehicles sampled from the trajectory dataset. To ensure the calibration accurately represents the general driving behaviors on the expressway, the sampling process is stratified by vehicle velocity. Specifically, 20 vehicles are sampled from five predefined velocity intervals each. The least squares method [45] is used to calibrated the model. The calibrated parameters are as follows: $v_0$ = 22.220 m/s; $a$ = 1.000 m/s$^2$; $b$ = 2.500 m/s$^2$; $s_0$ = 2.303 m; $T$ = 0.500 s. The validation results for vehicle location demonstrate a Root Mean Squared Error (RMSE) of 0.148 m and a Mean Absolute Percentage Error (MAPE) of 0.026%.

$$\dot{v}_{IDM} = a\left[1 - \left(\frac{v}{v_0}\right)^4 - \left(\frac{s^*}{s}\right)^2\right] \tag{5}$$

where $\dot{v}_{IDM}$ is the acceleration calculated based on the current velocity $v$ (m/s), desired speed $v_0$ (m/s), and current gap $s$ (m). The parameter to be calibrated is $a$ (the maximum acceleration). $s^*$ is the desired gap, defined as **Eq. (6).**

$$s^* = s_0 + vT + \frac{v\Delta v}{2\sqrt{ab}} \tag{6}$$

where $s^*$ is the desired gap. The parameters to be calibrated are $s_0$ (the minimum gap), $T$ (the desired time gap), and $b$ (the comfortable deceleration).

To apply the IVG vehicle grouping method, we need to determine three parameters, including sudden brake rate of leading vehicles, cut-in heading angle of adjacent vehicles, and response deceleration rate threshold of ego vehicles. The sudden brake rate of the leading vehicle is set to be the 90$^{th}$ percentile of all deceleration rates in the trajectory dataset, i.e., 2.00 m/s$^2$, while the response deceleration threshold of the ego vehicle is set to be the 80$^{th}$ percentile of all deceleration rates, i.e., 1.26 m/s$^2$, as shown in **Figure 4**. The cut-in heading angle of the adjacent vehicle is set to be the 85$^{th}$ percentile of all lane-change heading angles, i.e., 36.72°, as shown in **Figure 5**.

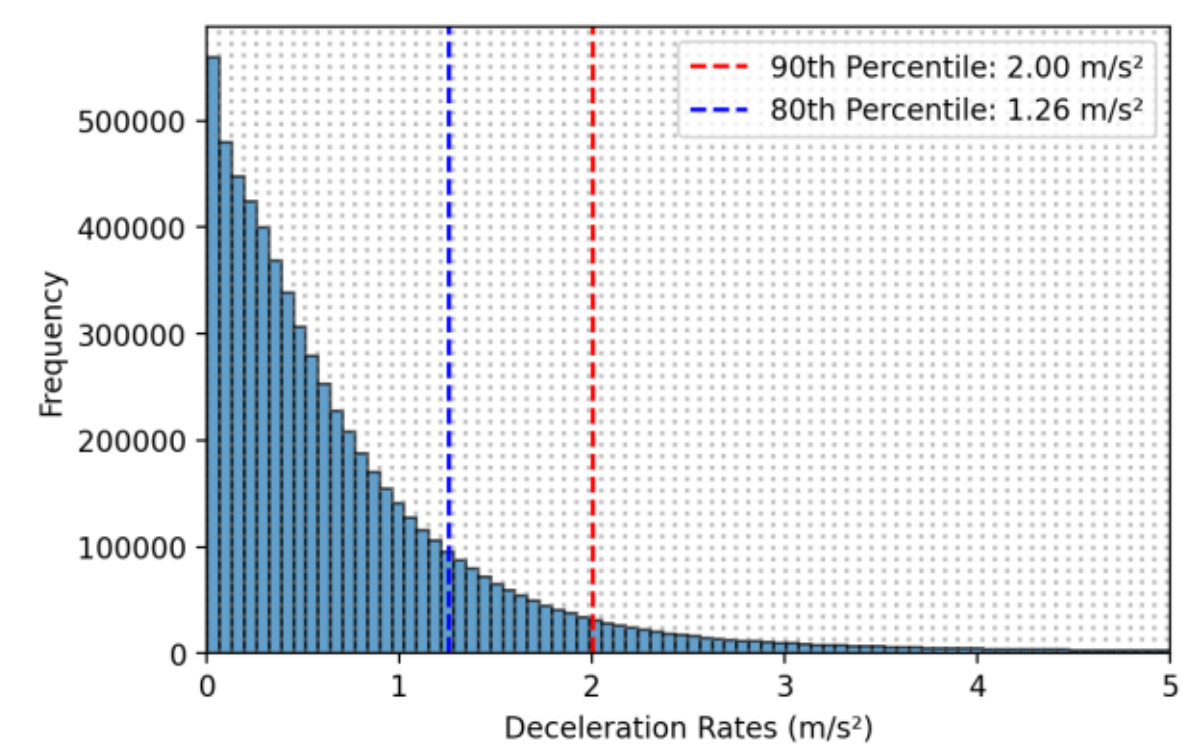

Fig. 4. Distribution of deceleration rates

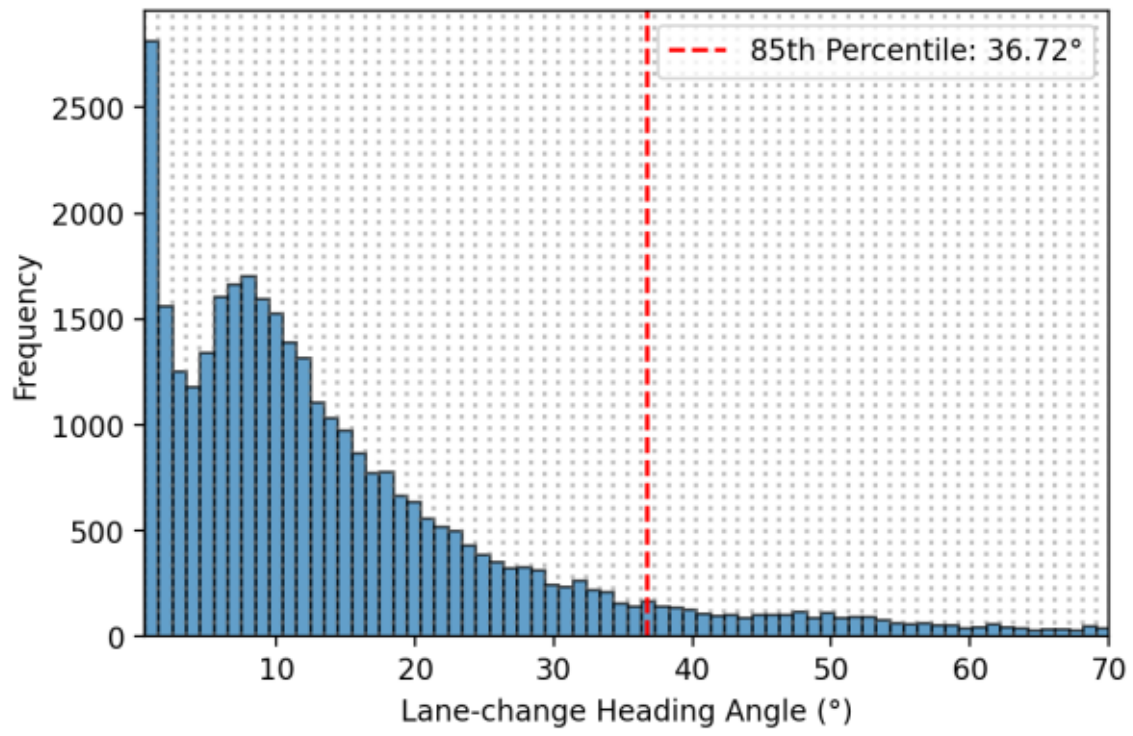

Fig. 5. Distribution of lane-change heading angles

## C. Data Preprocessing

After labeling the dependent variables based on the criteria in Section **II.B**, dataset imbalance is identified as a significant challenge. For VGRL, the ratio of Class 0 (no high risk) to Class 1 (high risk) exceeds 8:1, and a similar imbalance is observed for TRSC. This heavy imbalance could potentially impact prediction accuracy and the validity of result interpretation. To address this, a down-sampling method is employed to adjust the ratio to 4:1, consistent with practices in previous studies [46, 47]. The dataset is split into training and testing sets in a 70:30 ratio, which is a commonly used ratio by previous studies [12, 48]. To ensure consistent class distributions between the training and testing datasets [49], data records are first split by class before being assigned to the respective training and testing datasets.

For the LR model, after extracting aggregated information from VGs, the data is gone through outlier removal, feature discretization, and standardization to enhance the model's ability to identify underlying patterns. For the GNN model, all features and attributes are standardized prior to being input into the GNN to prevent the magnitude of the data from affecting the learning process.

# IV. Results

## A. LR using Aggregated VG Information

We employ LR to evaluate the impact of aggregated VG information on VG risks. The Area under the Receiver Operating Characteristic (ROC) curve (AUC) is utilized to assess the predictive ability of the models, where a higher AUC value indicates superior performance in prediction. The prediction results on VGRL and TRSC on the testing datasets are presented in **Figure 6**. Additionally, we conduct analyses using different prediction time intervals including 1s, 3s, and 5s. For example, a 1-second interval signifies using current information of a VG to predict VGRL and TRSC of the VG one second later. The results demonstrate that aggregated VG information enables LR to accurately predict VG risks, where the performance is enhanced as prediction intervals decrease. Specifically, the AUC values range from 0.89 to 0.94 for VGRL and from 0.81 to 0.86 for TRSC from 5s prediction interval to 1s prediction interval, indicating that more recent information is crucial for accurate VG risk prediction.

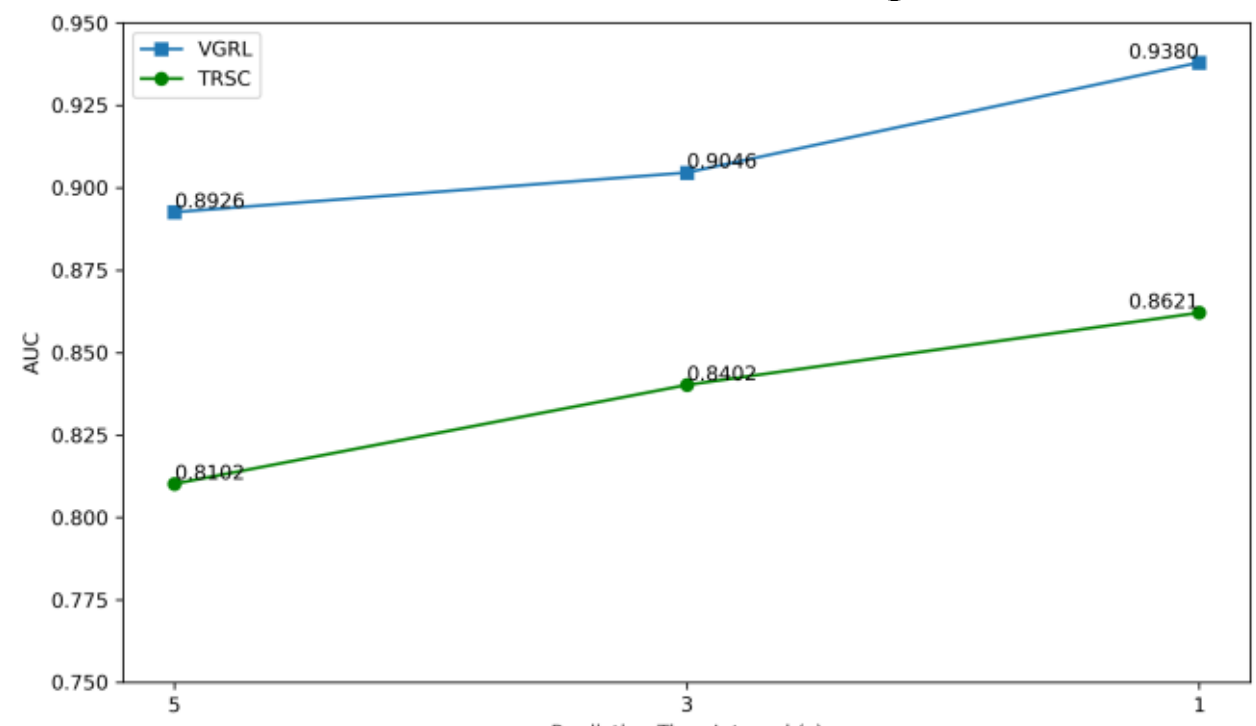

Fig. 6. Prediction performance using LR and aggregated VG information

TABLE III

COEFFICIENTS FOR LR MODEL WITH A 1-SECOND PREDICTION TIME INTERVAL

| Dependent Variable: VGRL (1- High Risk, 0 – Low Risk) | | | | |
|---|---|---|---|---|
| Variables | Coef. | Std. Error | Z value | P value |
| *avg_v* | -3.768 | 0.061 | -62.099 | <0.001 |
| *avg_a* | 0.112 | 0.042 | 2.683 | <0.001 |
| *std_v* | 1.047 | 0.419 | 2.499 | 0.007 |
| *pctg_large_veh* | 0.262 | 0.220 | 1.192 | 0.233 |
| *pctg_change_lane* | 3.959 | 0.107 | 37.117 | <0.001 |
| *risk_peak* | 3.662 | 0.040 | 91.128 | <0.001 |
| *num_on_ramp* | 2.241 | 0.212 | 10.592 | <0.001 |
| *num_off_ramp* | 2.790 | 0.207 | 13.457 | <0.001 |
| Dependent Variable: TRSC (1- Escalated, 0 – Not Escalated) | | | | |
| Variables | Coef. | Std. Error | Z value | P value |
| *avg_v* | -3.402 | 0.064 | -53.158 | <0.001 |
| *avg_a* | 0.494 | 0.046 | 10.641 | <0.001 |
| *std_v* | 1.275 | 0.052 | 24.685 | <0.001 |
| *pctg_large_veh* | 0.215 | 0.231 | 0.931 | 0.352 |
| *pctg_change_lane* | 2.616 | 0.119 | 21.965 | <0.001 |
| *risk_peak* | 1.377 | 0.043 | 32.263 | <0.001 |
| *num_on_ramp* | 0.763 | 0.213 | 3.580 | <0.001 |
| *num_off_ramp* | 1.252 | 0.206 | 6.071 | <0.001 |

**Table III** presents the coefficients from the LR models predicting VGRL and TRSC with a 1-second prediction time interval. The VG features exhibit similar effects on both outcomes: average velocity (*avg_v*) has a negative impact on risks, whereas the other variables have positive impacts. This pattern remains consistent for 3-second and 5-second prediction intervals. The negative impact of average velocity may seem counterintuitive, as speeding is a well-known factor for crashes [50]. However, a lower average speed signifies higher density and closer proximity among vehicles, thereby increasing the frequency of interactions and elevating risk levels and scales. This argument is supported by the capacity analysis of the MAGIC Dataset [44], where the traffic state is categorized into four types including smooth, relatively smooth, congested, and severely congested based on the Shanghai urban traffic state index. In most cases, states of the eastbound traffic (data used in this study) fall in the congested category, which results in the most intense vehicle interactions with increased risks of crash (e.g., rear end).

Additional key findings include the significant positive influence of a high percentage of lane-changing vehicles (*pctg_change_lane*) on VG risks. Lane-changing maneuvers are of high risk, as it requires matching the speed of nearby vehicles, identifying a suitable gap in traffic, ensuring that a driver's own movements are recognized by others, and a capable execution [51]. Failure in any of these aspects can result in insufficient gaps and significant speed discrepancies between vehicles, potentially causing abrupt braking and increasing VG risks. The standard deviation of vehicle speeds (*std_v*) within a VG also plays a crucial role, as greater variability in speeds reflects more complex vehicle behaviors, which correlate with higher risk level and broader risk scale of the VG.

The current Risk Indicator of a VG (*risk_indicator*) is another vital factor, suggesting that VGs already at high risk are likely to maintain this state. Furthermore, road geometry positively affects VG risks. Traffic merging from on-ramps can disrupt the main road flow, thereby increasing VG risks, while the diverging of vehicles to off-ramps necessitates lane changes and deceleration, which can also increase VG risks. These findings are consistent with previous studies [52, 53].

### *B. GNN using Disaggregated VG Information*

The disaggregated VG information, represented as graph data, can provide more insights explaining the development of VG risks. The GNN network configuration, prediction results, and interpretation are presented in this section.

#### *1) Network Configuration*

The network has 3 parametric layers at a single level of graph resolution. The network has 8 input channels of node features, 1 input channel of edge attribute, 128 hidden channels (same for three NNConv layers), and 1 output channel. Its configuration can be described as:

$$C(128) - ReLU - C(128) - ReLU - C(128) - ReLU - GMP - FC(1)$$

where $C(c)$ denotes an NNConv layer with $c$ output channels. $ReLU$ is ReLU activation. $GMP$ is global mean pooling. $FC(c)$ is a fully-connected layer with $c$ output channels.

Each NNConv layer $C(c)$ uses a separate filter-generating MLP, denoted $F^l$:

- $F^1$ (the first NNConv layer) has configuration: $FC(64) - ReLU - FC(128) - ReLU - FC(8 \times 128)$
- $F^3$ (the second NNConv layer) and $F^5$ (the third NNConv layer) each have configuration: $FC(64) - ReLU - FC(128) - ReLU - FC(256) - ReLU - FC(512) - ReLU - FC(128 \times 128)$

The network is trained with the Adam optimizer, utilizing a Binary Cross Entropy with Logits Loss function to handle the binary classification task.

#### *2) Prediction Performance*

As presented in **Figure 7**, with a 1-second prediction time interval, the GNN achieved an AUC exceeding 0.93 for VGRL and approximately 0.85 for TRSC. These results highlight the GNN's capability in predicting VG risks by leveraging disaggregated VG information. The superior performance in shorter time intervals remains evident. The lower performance in predicting TRSC is also consistent with previous observations in LR. However, it is important to acknowledge that, due to the complexity and inherent noise of graph data, the prediction performance of the GNN is slightly inferior to that of LR with aggregated data.

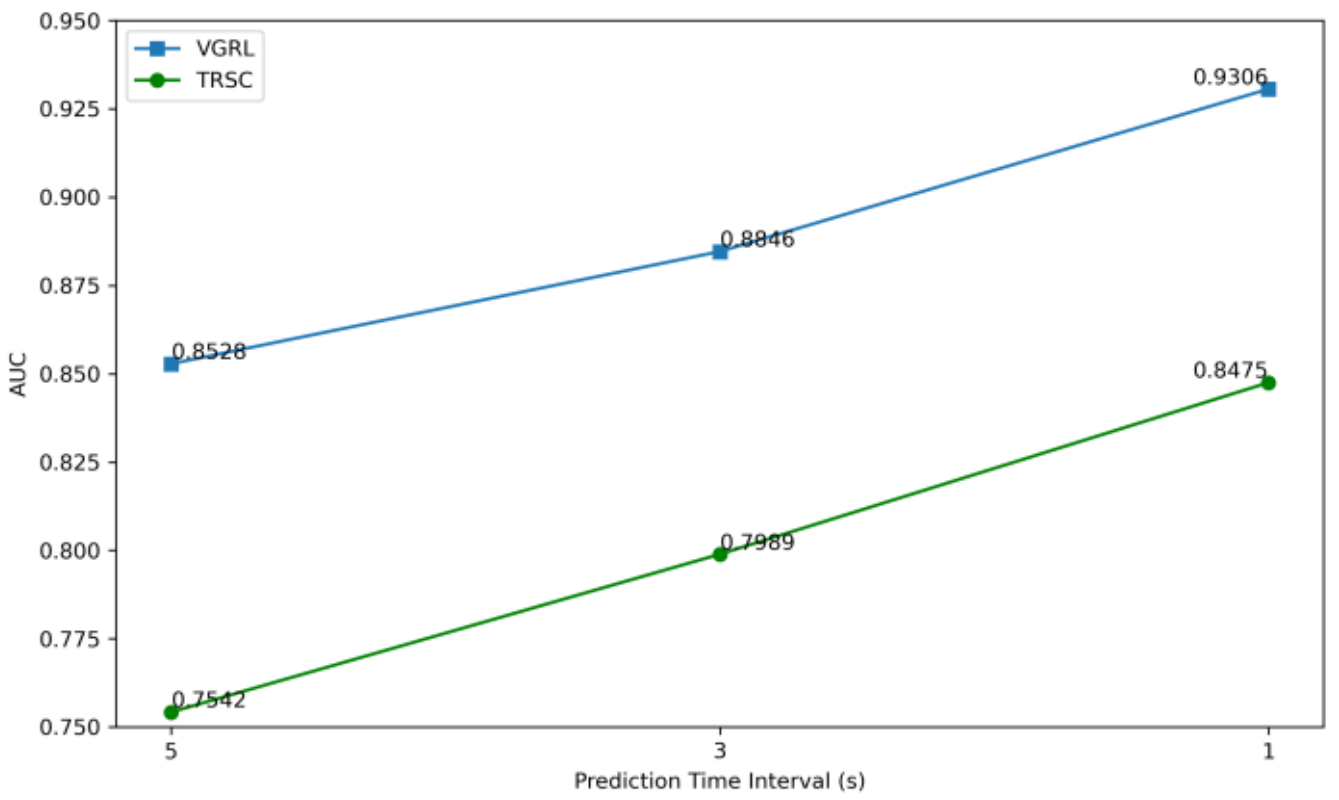

Fig. 7. Prediction performance using GNN and disaggregated VG data

#### *3) Result Interpretation*

The interpretation of the GNN results is showcased based on predictions with 1-second and 5-second time intervals. Two node features, namely *lane_id* and *dist*, are excluded from this interpretation as they primarily serve to indicate node locations within the graph and do not carry substantial semantic information.

The following steps describes how the data preparation is conducted:

- Identify significant node features: we employ GNNExplainer to generate graph masks for all graphs in the testing dataset. We consider significant node features as whose mask values exceeding 90$^{th}$ percentile of all values across all graphs. The following analysis is based on those significant node features.
- Feature perturbation analysis: to assess the directional impact of these significant node features, feature perturbation is performed. After loading the trained GNN model, we iteratively set each significant node feature to zero and measure the resultant change in the model's output, referred to as delta output. Given that all graph data are standardized, zeroing out a node feature effectively reduces its magnitude. A negative delta output indicates that the node feature positively influences the output, whereas a positive delta output signifies a negative influence.

- Logging significant features: Each significant node feature is logged with its feature type, feature value, node location, and delta output for subsequent analysis.

**Figure 8** illustrates the distribution of feature types among the significant node features. The distributions for VGRL and TRSC predictions are consistent, with vehicle velocity (*v*), lane change (*changing_lane*), and acceleration (*abs_a*) emerge as the three dominant feature types. Notably, vehicle velocity is the most critical feature in assessing VG risk, accounting for over half of the significant feature types in both prediction models. The observation is consistent across all prediction time intervals.

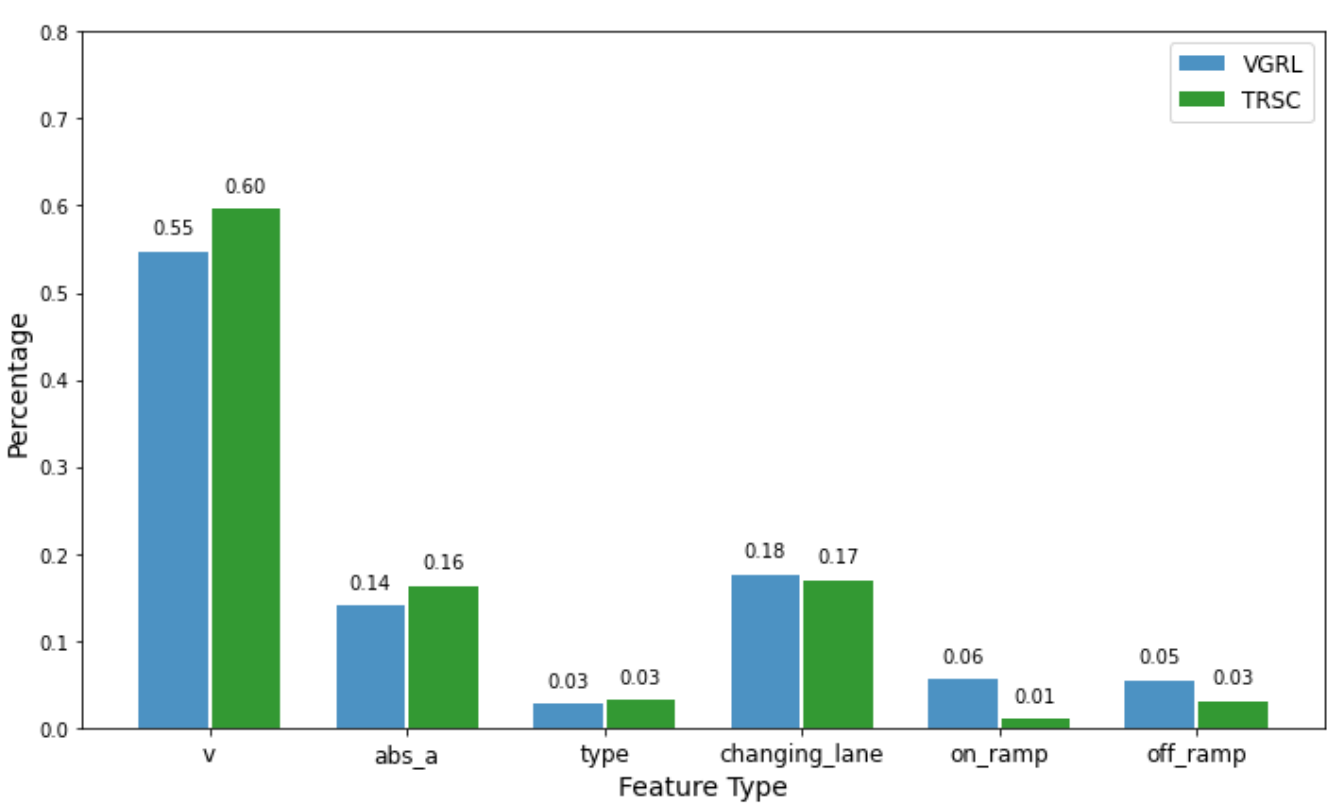

(a) 1-second prediction time interval

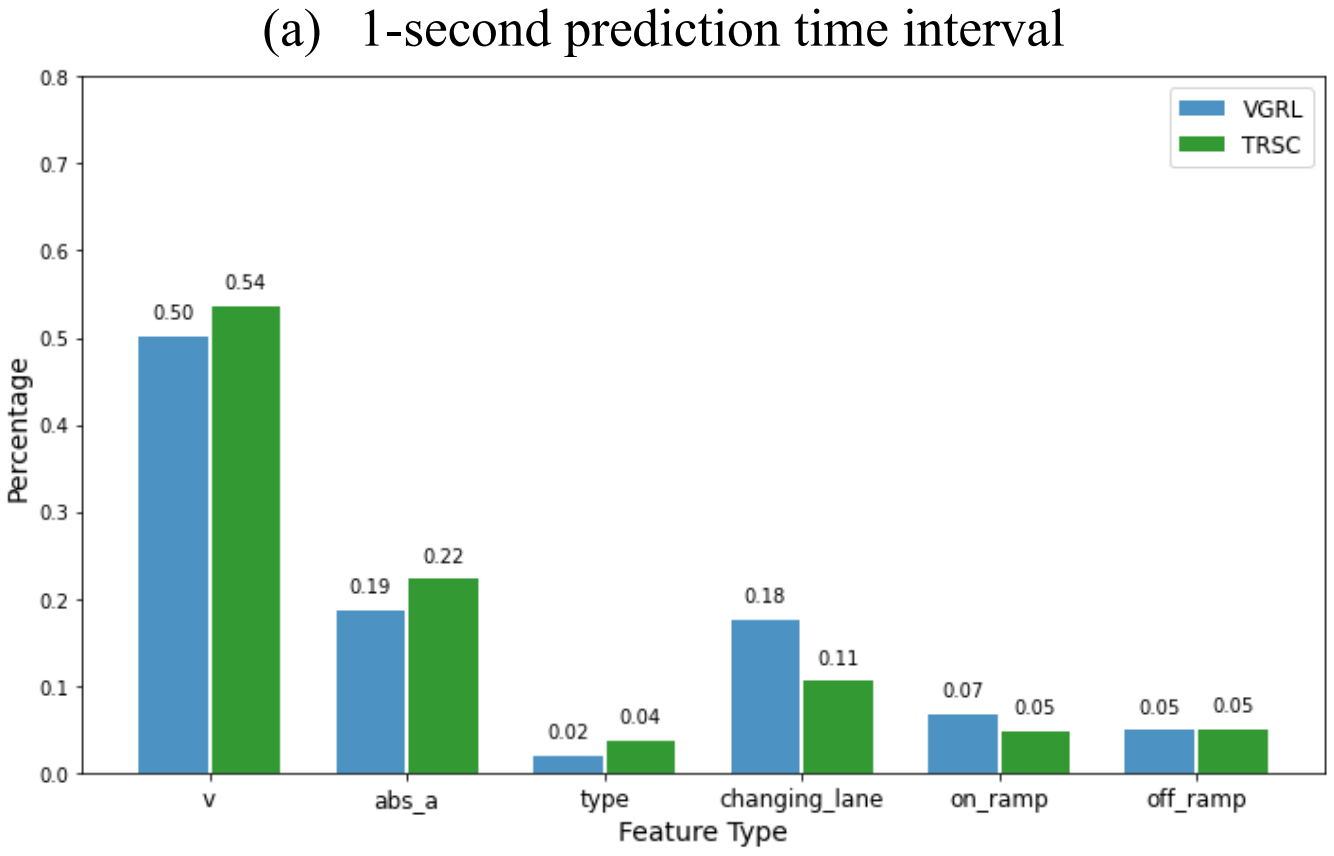

(b) 5-second prediction time interval

Fig. 8. Distribution of feature types of significant node features

**Figure 9** and **Figure 10** present the distribution of delta output values of significant node features using 1s and 5s prediction time interval, depicted by violin plots, which shows the general directional impact of different feature types. The results indicate a clear trend for both VGRL and TRSC predictions: velocity (*v*) exerts a negative impact on the output, while other feature types tend to have a positive impact. This finding aligns with the results obtained from the LR models utilizing aggregated VG information. The directional impact of feature types is also similar using different prediction time intervals.

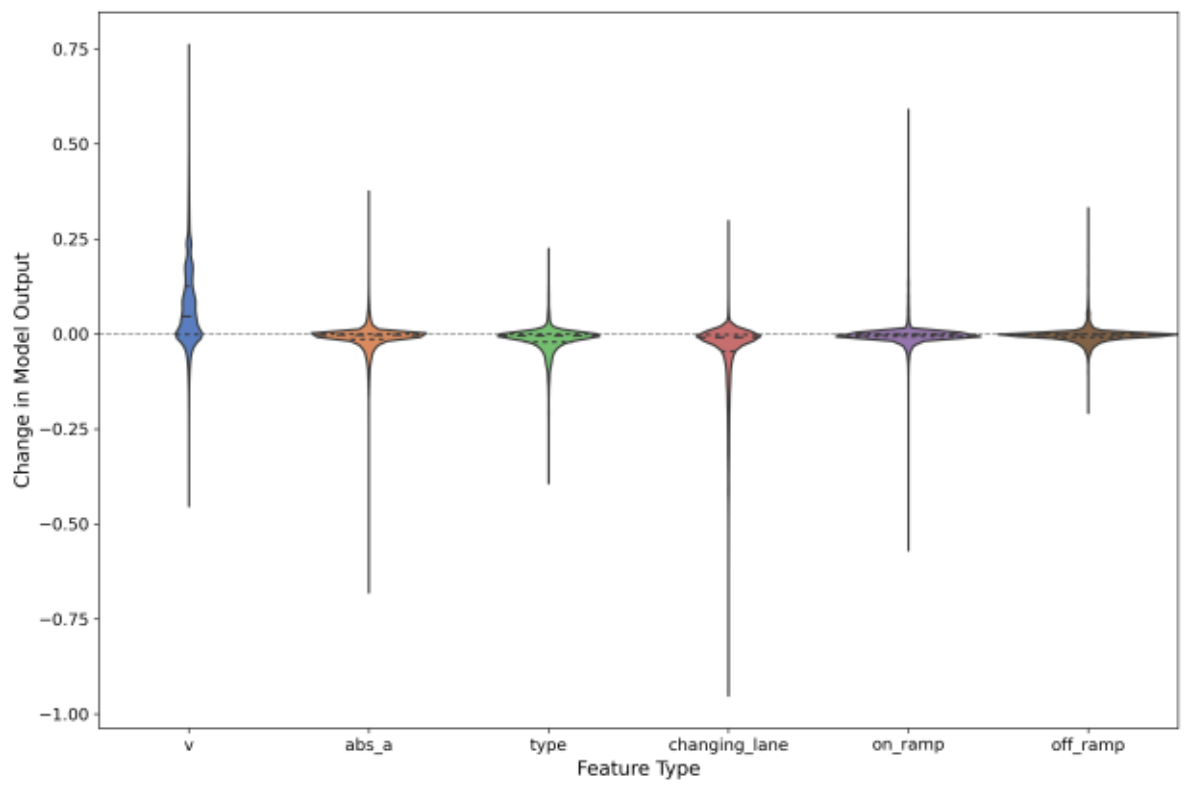

(a) VGRL

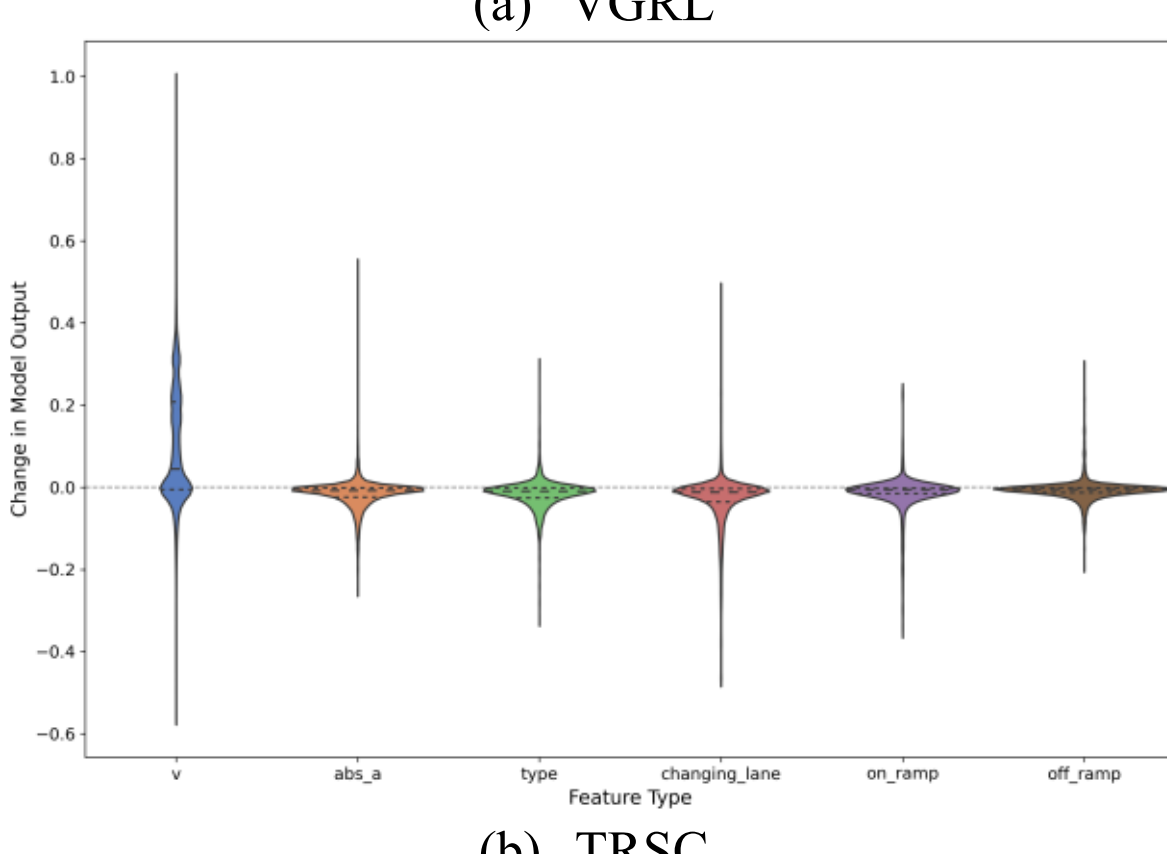

(b) TRSC

Fig. 9. Violin plots of delta output (1s)

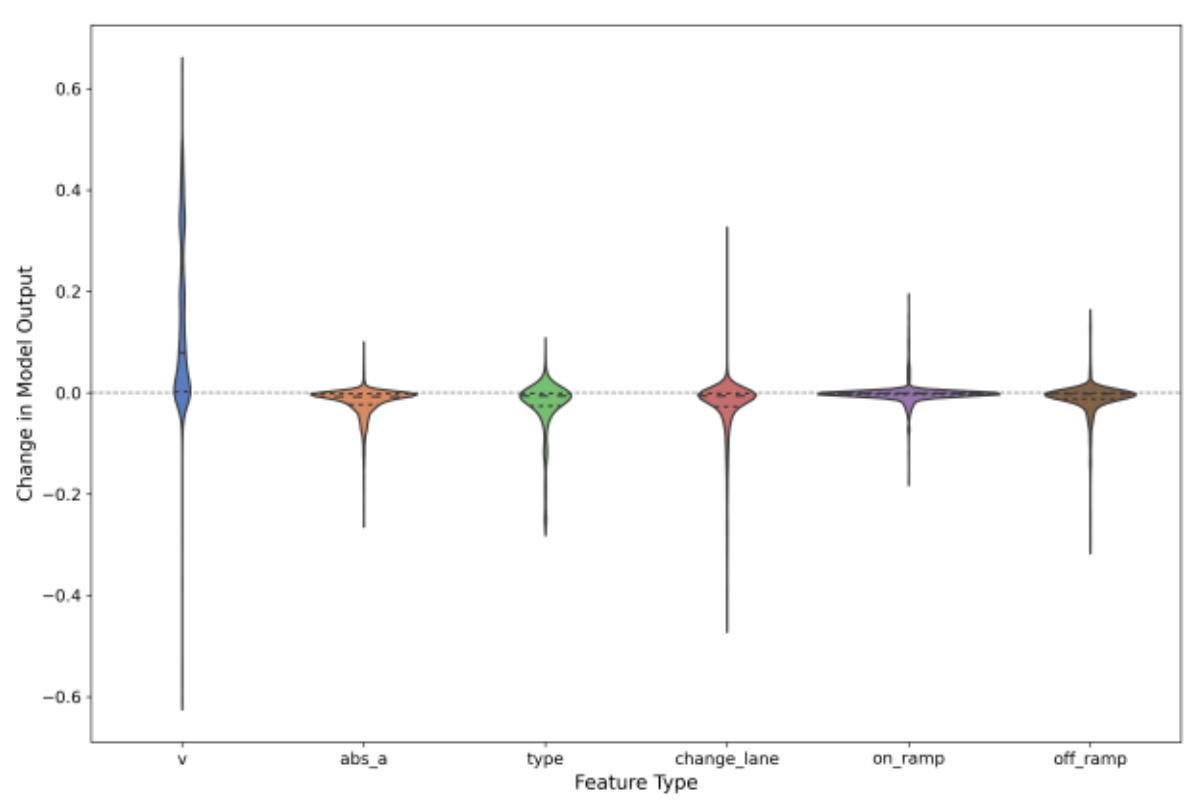

(a) VGRL

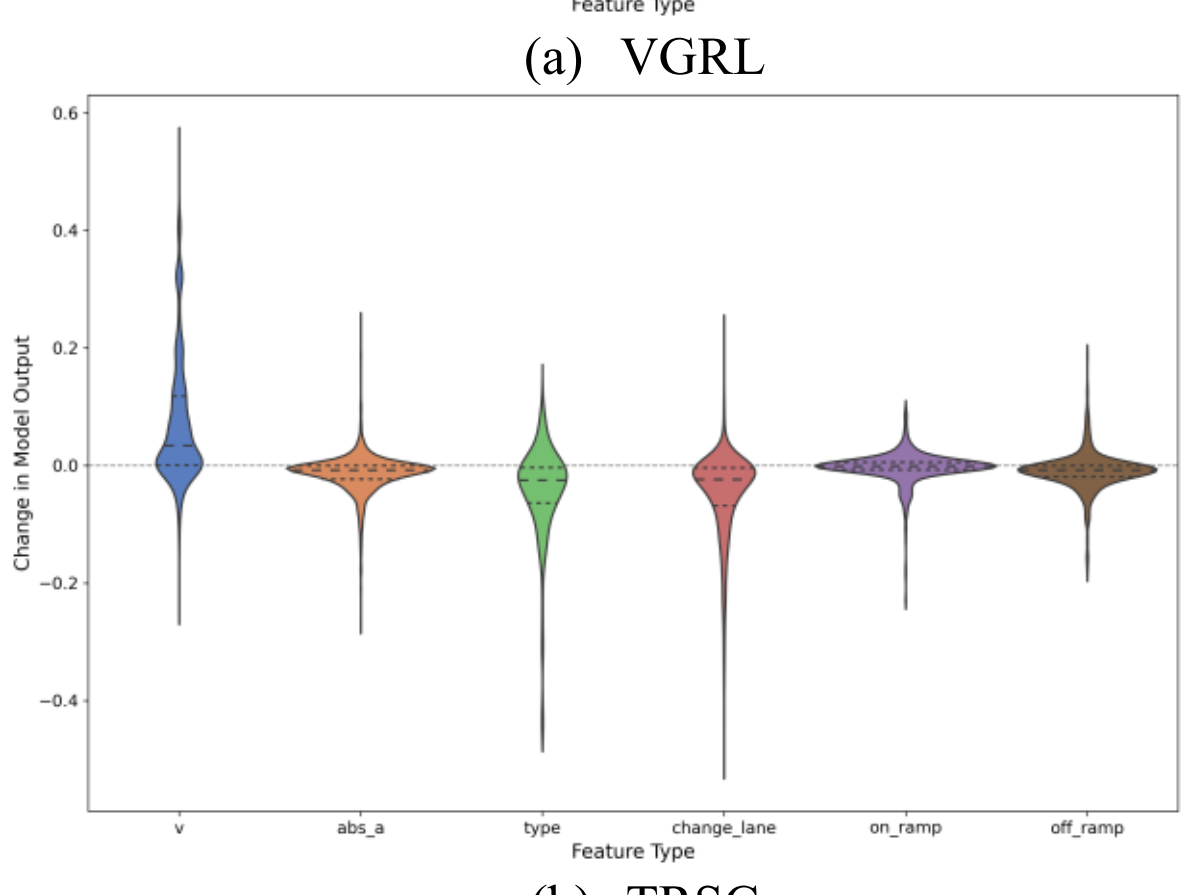

(b) TRSC

Fig. 10. Violin plots of delta output (5s)

To explore the spatial distributions of significant node features within VGs, we focus on features belonging to the three dominant types that exhibited consistent directional impacts including velocity (*v*), lane change (*change_lane*) and acceleration (*abs_a*).

We further refine the selection of significant node features by focusing exclusively on those associated with medium-sized VGs, defined as groups comprising 8 to 15 vehicles, because (1) smaller VGs often lack sufficient data to reveal meaningful spatial patterns of significant node features; and (2) larger VGs tend to introduce noise in the spatial distribution of significant node features, making interpretation challenging.

**Figure 11** and **Figure 12** present the spatial distribution of significant node features using 1s and 5s prediction time intervals in heatmaps, respectively. In each heatmap, the y-axis represents the longitudinal positions of nodes within a VG (i.e., head or tail). The x-axis denotes the lateral positions of nodes (i.e., left or right lanes).

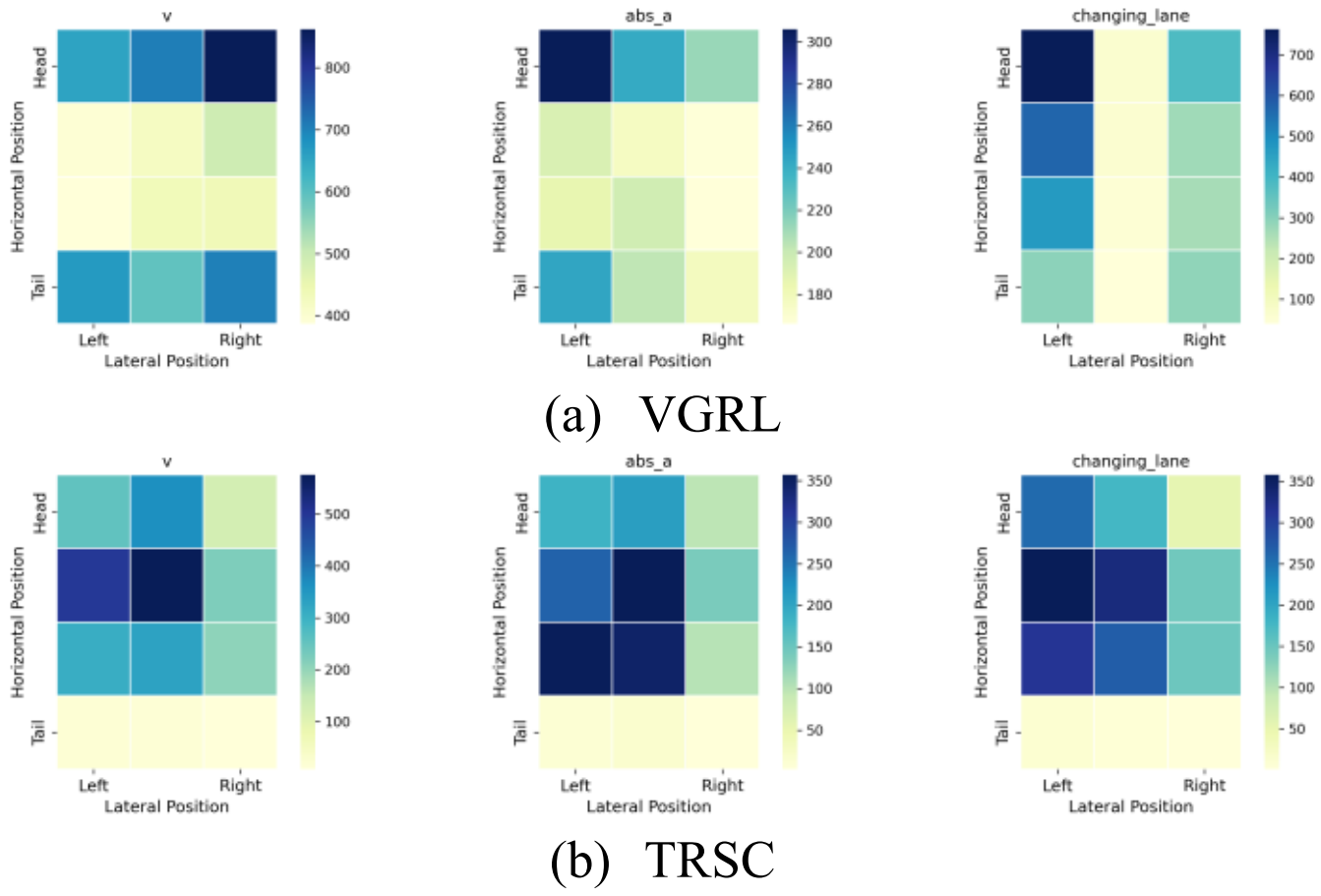

(a) VGRL

(b) TRSC

Fig. 11. Spatial distribution of significant node features (1s)

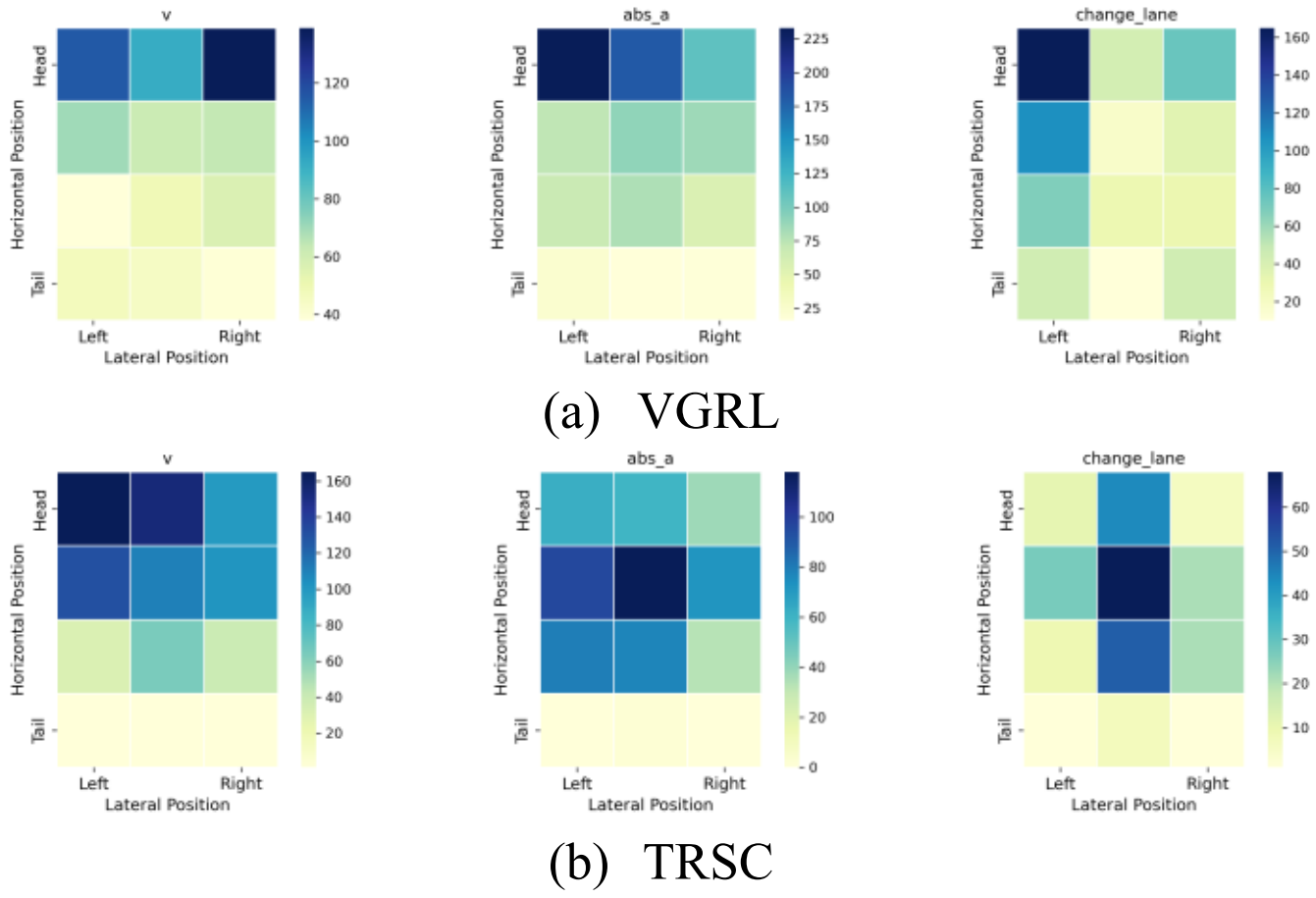

(a) VGRL

(b) TRSC

Fig. 12. Spatial distribution of significant node features (5s)

A clear pattern is observed in VGRL prediction: significant node features concentrate close to the head of VGs. This suggests that behaviors of leading vehicles (i.e., speed, acceleration, lane-changing) contribute to the elevation of risk level of a VG. This is because the behavior of leading vehicles can trigger cascading responses from following vehicles, which may be amplified and causes more disruptive traffic flow [54].

However, for TRSC prediction, the significant node features are primarily distributed toward the center of VGs. This suggests that slow-moving vehicles, drastic speed changes, and lane changes occurring in the central positions of VGs are more likely to escalate the Risk Scale, possibly due to their interactions with multiple surrounding vehicles within the group. These interactions can lead to compounded effects, where a single central vehicle's abrupt behavior influences several adjacent vehicles simultaneously, thereby magnifying the Risk Scale.

The main findings of this section can be concluded as:

(1) Lower speed in VGs could increase crash risks due to more frequent vehicle interactions under congested traffic states.
(2) Abrupt changes in speed, lane changes, and proximity to ramps could intensify VG risks, aligning with previous studies.
(3) VGs tend to remain at high risk once established.
(4) The adverse maneuvers, such as lane changes and sudden changes in speed, in the front of a VG could result in higher VG risk levels.
(5) The adverse maneuvers close to the center of a VG often contribute to an increase in the VG's Risk Scale.

## V. Discussion and Conclusions

Previous studies on predicting crash risks for expressways primarily associated crash occurrences with traffic parameters or the geometric features of road segments. However, these studies usually overlooked the impact of vehicles' continuous movement and interactions with surrounding vehicles on crash risks. By leveraging high-resolution vehicle trajectory data, we introduced the concept of vehicle groups (VGs) to capture the impact of vehicle interactions. We proposed an impact-based vehicle grouping (IVG) method to cluster vehicles into groups, and investigated how current VG states influence risks in future time steps.

A Logistic Regression (LR) and a Graph Neural Network (GNN) were employed to predict VG risks, using aggregated and disaggregated information respectively. The performance, measured by the AUC, exceeded 0.93 for predicting the risk levels of VGs and reached 0.86 for the prediction of changes in Risk Scale in 1-second prediction time interval. Furthermore, we introduced an explainable framework that integrates GNNExplainer with feature perturbation to interpret the GNN outcomes.

The methodologies and findings of this study offer valuable insights in designing a Proactive Traffic Safety Management (PTSM) system. Specifically:

(1) The IVG method delineates clusters of interacting vehicles as the target for safety interventions.
(2) The prediction models accurately assess changes in both the risk level and risk scale for VGs.
(3) The explainable GNN framework can be beneficial to detection and mitigation of the adverse vehicle maneuvers within the VGs.

The main limitation of this study lies in the selection of risk indicators. Integrating factors of more types, such as driver habits, weather, and road surface conditions, could further enrich the analysis and yield more robust insights.

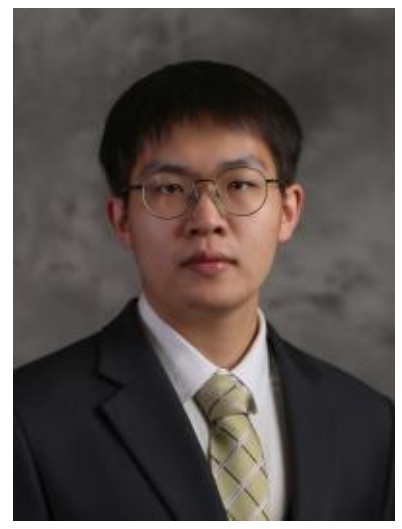

**Tianheng Zhu** received the B.S. degree in transportation engineering from Tongji University, Shanghai, China, in 2023. He is currently pursuing the Ph.D. degree and working as a Graduate Research Assistant with the Lyles School of Civil and Construction Engineering, Purdue University. His current research interests include traffic safety and traffic operations and control.

Mr. Zhu has published several papers at top journals and conferences like Transportation Research Part C, IEEE ITSC and TRB Annual Meeting.

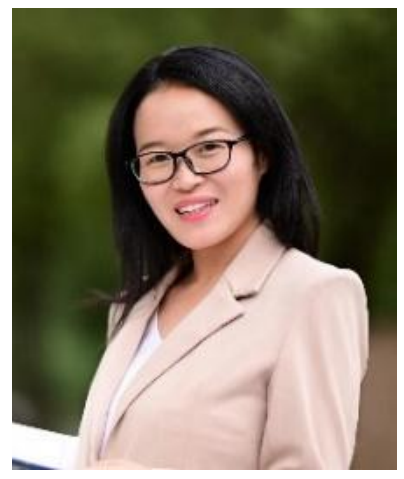

**Ling Wang** received the Ph.D. degree in transportation engineering from the University of Central Florida in 2016.

She is currently an Associate Professor at Tongji University, Shanghai, China. Her research interests include traffic safety, active traffic management, and big data analytics.

Prof. Wang has published more than 40 papers in top journals. She has received the Best Paper Award of World Transportation Congress and the Best Young Researcher Award of TRB as the first author.

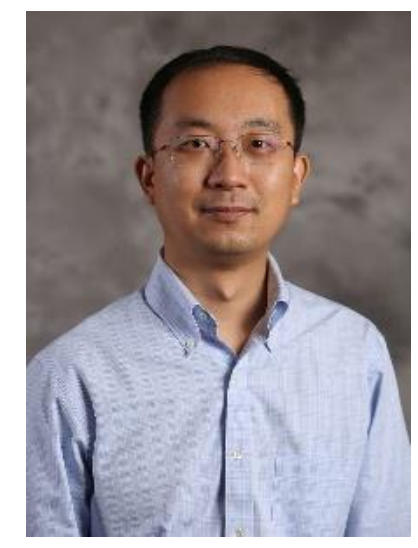

**Yiheng Feng** (Member, IEEE) received the B.S. and M.E. degrees from the Department of Control Science and Engineering, Zhejiang University, Hangzhou, China, in 2005 and 2007, respectively, and the Ph.D. degree in systems and industrial engineering from the University of Arizona in 2015.

He is currently an Assistant Professor with the Lyles School of Civil and Construction Engineering, Purdue University. His research interests include traffic operations and control, cybersecurity of the transportation systems and connected and automated vehicle testing and evaluation.

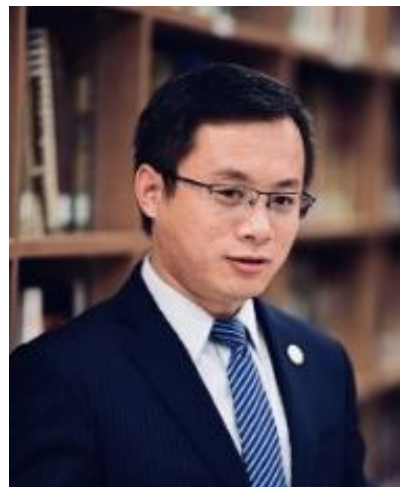

**Wanjing Ma** received the Ph.D. degree in traffic engineering from Tongji University, Shanghai, China, in 2007.

He is currently a Professor and the Head of the College of Transportation Engineering, Tongji University. His research interests include traffic operation and control, intelligent transportation systems, and shared mobility.

Prof. Ma has published more than 100 articles in top journals. He was awarded the title of Elsevier Highly Cited Researcher in China in 2021.

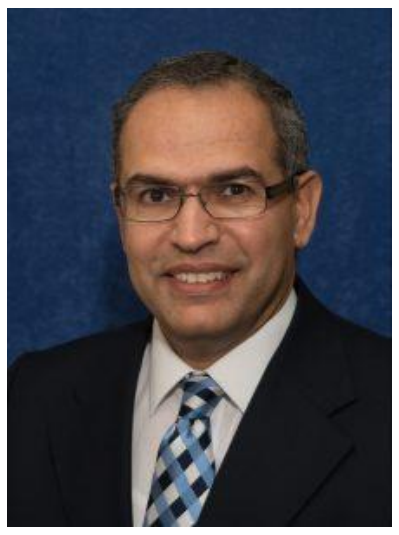

**Mohamed Abdel-Aty** (Senior Member, IEEE) received the Ph.D. degree in civil engineering from the University of California, Davis, in 1995.

He is currently a Pegasus Professor and the Chair of the Civil, Environmental, and Construction Engineering Department, University of Central Florida, Orlando, FL, USA. His main expertise and interests are in the areas of ITS, simulation, CAV, and active traffic management.

Prof. Abdel-Aty has managed over 75 research projects. He has published more than 750 papers, more than 400 in journals (As of August 2023, Google Scholar citations: 29741, H-index: 95). He received nine best paper awards from ASCE, TRB, and WCTR. He is the Editor-in-Chief Emeritus of Accid. Anal. Prev.